\documentclass[conference]{IEEEtran}
\IEEEoverridecommandlockouts

\usepackage{cite}
\usepackage{amsmath,amssymb,amsfonts}
\usepackage{algorithmic}
\usepackage{graphicx}
\usepackage{textcomp}
\usepackage{xcolor}
\usepackage{color}
\usepackage{tabularray}
\usepackage{pifont}
\usepackage{url}
\usepackage{booktabs}
\usepackage{multirow}
\usepackage{balance}
\usepackage{colortbl}

\def\BibTeX{{\rm B\kern-.05em{\sc i\kern-.025em b}\kern-.08em
    T\kern-.1667em\lower.7ex\hbox{E}\kern-.125emX}}
\begin{document}
\newcommand{\cmark}{\textcolor{green!60!black}{\ding{51}}}  
\newcommand{\xmark}{\textcolor{red}{\ding{55}}}             

\title{Bidirectional Mammogram View Translation with Column-Aware and Implicit 3D Conditional Diffusion}
\author{\IEEEauthorblockN{Xin Li$^{\dag}$, Kaixiang Yang$^{\dag}$, Qiang Li, Zhiwei Wang$^{*}$ }
\IEEEauthorblockA{\textit{Wuhan National Laboratory for Optoelectronics, Huazhong University of Science and Technology} \\
\textit{$^\dag$: Co-first authors, $^*$: Corresponding authors.}\\
zwwang@hust.edu.cn}
}

\maketitle

\begin{abstract}
Dual-view mammography, including craniocaudal (CC) and mediolateral oblique (MLO) projections, offers complementary anatomical views crucial for breast cancer diagnosis. However, in real-world clinical workflows, one view may be missing, corrupted, or degraded due to acquisition errors or compression artifacts, limiting the effectiveness of downstream analysis.
View-to-view translation can help recover missing views and improve lesion alignment. Unlike natural images, this task in mammography is highly challenging due to large non-rigid deformations and severe tissue overlap in X-ray projections, which obscure pixel-level correspondences.
In this paper, we propose Column-Aware and Implicit 3D Diffusion (CA3D-Diff), a novel bidirectional mammogram view translation framework based on conditional diffusion model. 
To address cross-view structural misalignment, we first design a column-aware cross-attention mechanism that leverages the geometric property that anatomically corresponding regions tend to lie in similar column positions across views. A Gaussian-decayed bias is applied to emphasize local column-wise correlations while suppressing distant mismatches.
Furthermore, we introduce an implicit 3D structure reconstruction module that back-projects noisy 2D latents into a coarse 3D feature volume based on breast-view projection geometry. The reconstructed 3D structure is refined and injected into the denoising UNet to guide cross-view generation with enhanced anatomical awareness.
Extensive experiments demonstrate that CA3D-Diff achieves superior performance in bidirectional tasks, outperforming state-of-the-art methods in visual fidelity and structural consistency.
Furthermore, the synthesized views effectively improve single-view malignancy classification in screening settings, demonstrating the practical value of our method in real-world diagnostics.
These advancements position CA3D-Diff as a promising tool for clinical applications, particularly in missing view recovery and improved cross-view representation learning.
Our code is available at \url{https://github.com/lixinHUST/CA3D-Diff}.
\end{abstract}


\section{Introduction}
Breast cancer is the most common malignancy among women worldwide~\cite{hosseini2016early}, with mammography playing a pivotal role in its early detection and diagnosis~\cite{moss2012impact}. 
In a standard mammographic screening examination, the 3D breast is projected onto 2D X-ray images under two typical views: cranio-caudal (CC), where the X-ray penetrates from top to bottom, and mediolateral oblique (MLO), where the X-ray is angled at approximately 45 degrees from the upper inner part of the breast. This dual-view imaging is both necessary and sufficient for radiologists to comprehensively assess breast anatomy and detect potential abnormalities.
This dual-view paradigm also serves as the foundation for modern computer-aided diagnosis (CAD) systems~\cite{dchanet,li2021deep,petrini2022breast}, which aim to enhance detection accuracy and reduce diagnostic workload by leveraging deep learning techniques.

There is growing interest in bidirectional mammogram view translation, which involves synthesizing one standard view (\textit{e.g.}, MLO) from the other (\textit{\textit{e.g.}}, CC), or vice versa. This task holds significant clinical value, enabling recovery of missing views, improved lesion correspondence analysis, enhanced data augmentation, and better cross-view representation learning. However, it remains highly challenging due to the substantial structural differences between views. These differences arise from variations in projection angle, breast compression, and tissue deformation. More critically, dense tissue overlap caused by X-ray projection severely obscures spatial correspondence, making cross-view mapping far more ambiguous than in natural image domains. As the two views are fundamentally distinct 2D projections of a complex 3D anatomy, precise pixel-wise alignment is often unattainable. Moreover, the cross-view mapping is nonlinear, spatially non-rigid, and lacks direct pixel-level supervision.


View-to-view translation can be regarded as a specific form of image-to-image translation. Existing methods predominantly fall into two categories: generative adversarial networks (GANs)~\cite{GAN} and denoising diffusion probabilistic models (DDPMs)~\cite{DDPM}. 
GAN-based methods~\cite{pix2pix,pix2pixhd,cyclegan} typically adopt a UNet generator to map input images to target domains, paired with a PatchGAN discriminator for adversarial supervision. However, they often suffer from unstable training~\cite{gansurvey} and may introduce artifacts in clinically critical regions.
In contrast, diffusion-based approaches~\cite{img2img-turbo,sdedit,dit} model image generation as an iterative denoising process from Gaussian noise, offering greater training stability and improved image fidelity. Recent advances have further extended their applicability to conditional and guided synthesis. Moreover, 3D-aware diffusion models~\cite{zero123,syncdreamer} have emerged, learning implicit projection priors to synthesize novel views, and demonstrating strong generalization to unseen perspectives.

Despite these advances, existing methods are often developed for natural images and lack domain-specific priors required for medical images. A few methods~\cite{resvit,syndiff,medgan,GCDM} have explored image synthesis in the medical domain, yet they typically target modality translation (\textit{e.g.}, MRI-to-CT) and do not consider geometric relationships across views. In mammography, view translation is fundamentally different from modality translation, which involves a shared underlying anatomy captured from different 2D projections of a 3D structure.

In this paper, we propose Column-Aware and Implicit 3D Diffusion (CA3D-Diff), a novel conditional diffusion framework for bidirectional mammogram view translation. To address the inherent structural misalignment between CC and MLO views, CA3D-Diff introduces a column-aware cross-attention mechanism that leverages the anatomical prior that corresponding tissues often align along vertical axes across views. Additionally, we incorporate an implicit 3D modeling strategy by back-projecting 2D latents into a shared 3D feature space based on known breast projection geometry, enabling the model to capture cross-view anatomical consistency. Our unified framework supports both CC-to-MLO and MLO-to-CC synthesis with a single model, facilitating efficient learning of shared representations and improving generation fidelity across directions.

Our key contributions are summarized as follows:
\begin{itemize}
\item We propose CA3D-Diff, a novel conditional diffusion framework for bidirectional mammogram view translation, which addresses the inherent challenges of non-rigid deformation and pixel-wise misalignment between mammographic views.
\item We design a column-aware cross-attention mechanism that incorporates anatomical priors via column-wise locality and a Gaussian-decayed positional bias to improve structural consistency.
\item We introduce an implicit 3D structure reconstruction module that lifts 2D noisy latents into a coarse 3D feature volume based on projection geometry, providing anatomically grounded generation guidance.
\item We conduct comprehensive experiments on the VinDr-Mammo dataset~\cite{dataset}, demonstrating that CA3D-Diff outperforms state-of-the-art methods in terms of visual fidelity and structural consistency.
\item We validate the clinical utility of our method through a downstream screening task, where synthesized views effectively improve single-view malignancy classification performance.

\end{itemize}

\begin{figure*}[ht]
    \centering
    \includegraphics[width=0.9\linewidth]{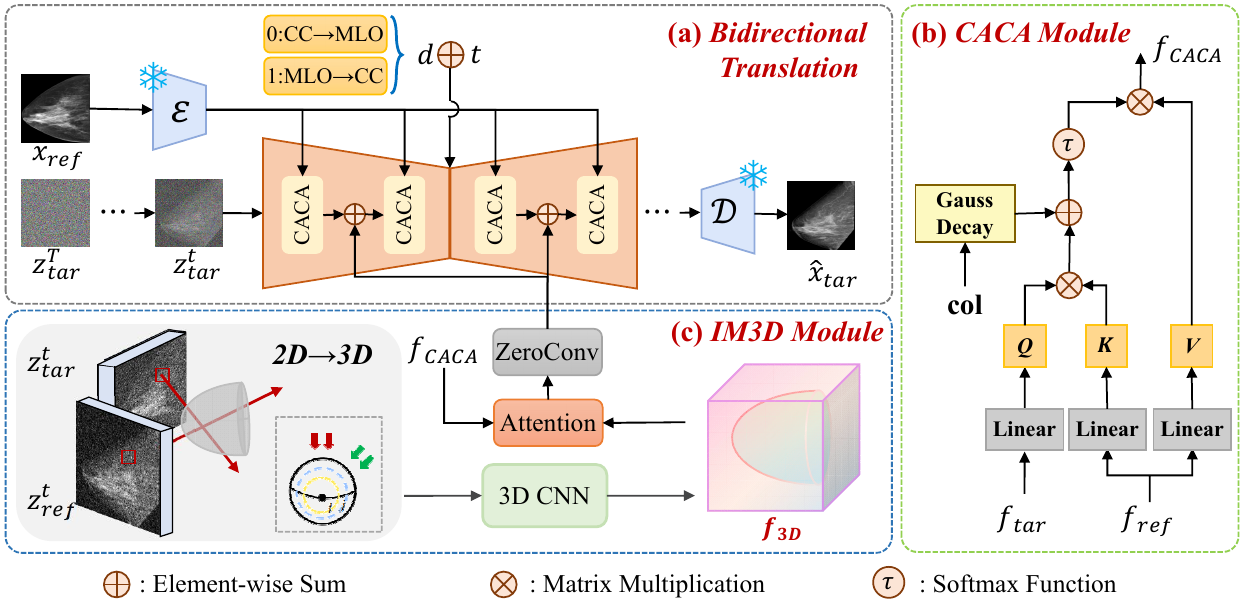}
    \caption{Overview of the proposed CA3D-Diff framework for bidirectional mammogram view translation.
}
    \label{fig:1}
\end{figure*}

\section{Related works}
\subsection{GAN-based Image Translation}

Generative Adversarial Networks (GANs)~\cite{GAN} have long served as the foundation for image-to-image translation. In the paired setting, Pix2pix~\cite{pix2pix} and its high-resolution variant Pix2PixHD~\cite{pix2pixhd} learn a conditional mapping from source to target images using a UNet generator and a PatchGAN discriminator, achieving fine-grained synthesis guided by pixel-wise supervision. For unpaired translation, CycleGAN~\cite{cyclegan} introduces a cycle-consistency loss to enforce mutual mappings between two domains without requiring aligned data. These models have inspired a broad range of style and modality translation methods in both natural and medical imaging. 

However, GAN-based methods often suffer from training instability and mode collapse~\cite{gansurvey}, they tend to  prioritize visual realism over structural correctness. 
Our task focuses on anatomically consistent translation between mammographic views, where each view exhibits large non-rigid deformations and complex spatial correspondences. Conventional GAN-based methods are ill-suited for this scenario due to their limited ability to model such structural complexity.

\subsection{DDPM-based Image Translation}
Denoising diffusion probabilistic models (DDPMs)\cite{DDPM} have emerged as a powerful alternative to GANs for image translation, offering stable training and high sample quality. In DDPMs, a sample is generated by gradually denoising a Gaussian noise vector using a learned reverse diffusion process. Extensions such as DDIM\cite{DDIM} and SDEdit~\cite{sdedit} further improve sampling efficiency and editing capability by enabling non-Markovian or guided sampling.

To improve generation speed and efficiency, recent advances have proposed various architectural and algorithmic enhancements. Img2img-Turbo\cite{img2img-turbo} introduces a single-step inference framework that accelerates conditional diffusion-based image editing, enabling real-time high-resolution generation. DiT\cite{dit} replaces conventional UNet backbones with pure transformer-based~\cite{transformer} architectures, enhancing global context modeling and improving generation scalability across image resolutions and domains.

Beyond image generation in 2D space, several works have explored integrating 3D awareness into diffusion models via camera-conditioned synthesis. Zero123\cite{zero123} introduces a 3D-aware diffusion framework that synthesizes novel views of an object from a single image, leveraging implicit geometry and known camera poses. Similarly, SyncDreamer\cite{syncdreamer} extends this idea to generate synchronized multi-view images using view frustum volume, enabling consistent geometry across diverse viewpoints. These methods model the image formation process as a projection from a latent 3D space, enhancing structural coherence in generated views.
While they are primarily developed for natural images and rely on known camera geometry or object-centric assumptions, their view-synthesis paradigm is highly relevant to mammographic image generation, where CC and MLO views can be understood as 2D projections from a latent 3D breast volume. 

\subsection{Medical Image Translation}
Medical image translation has received increasing attention due to its potential in cross-modality synthesis, artifact correction, and missing modality imputation. Compared to natural image translation, medical scenarios demand much higher fidelity and structural consistency, especially when synthesizing anatomically precise outputs.

GAN-based methods have been widely adopted in early works. MedGAN~\cite{medgan} introduces an end-to-end framework combining adversarial losses with perceptual and style-transfer losses for PET-CT translation. ResViT~\cite{resvit} integrates CNNs with residual vision transformer (ViT)~\cite{vit} blocks to capture long-range dependencies for MR-to-CT translation. 
Recently, diffusion models have shown promising performance in medical synthesis. CDM\cite{cdm} proposes a modality-specific representation and decoupled diffusion network to efficiently generate missing MRI sequences with high-quality outputs. Similarly, SynDiff\cite{syndiff} introduces an adversarial diffusion framework that combines conditional DDPMs with a cycle-consistent architecture, which enables translation between unpaired modalities by learning direct noise-to-image mappings while maintaining distribution consistency.

In the context of cross-view synthesis, our task differs significantly from previous modality translation settings. Tsuji et al.~\cite{mlo2cc} applied CR-GAN~\cite{crgan} for MLO-to-CC translation but did not incorporate breast-specific anatomical designs, leading to challenges in maintaining anatomical consistency.
We focus on bidirectional generation between two 2D mammographic views, which are projections from different angles of the same 3D breast anatomy. We draw inspiration from DiffuX2CT~\cite{diffux2ct}, which proposes a 3D-aware diffusion model for reconstructing volumetric CT from 2D biplanar X-rays. In our method, we uniquely incorporate both projection prior and column-wise anatomical correlations. Compared to DCHA-Net~\cite{dchanet}, which also emphasizes dual-view correlation for classification, our method directly integrates geometric priors into the image synthesis process through column-aware cross-attention and implicit 3D structure reconstruction modules, achieving anatomically consistent and clinically meaningful view translation.

\section{Methods}
\subsection{Bidirectional View Translation with Conditional Diffusion}
We propose a latent diffusion-based framework for bidirectional mammogram view translation between CC and MLO projections, as illustrated in Fig.~\ref{fig:1}a. Trained on paired CC-MLO data, the framework enables controllable synthesis in both directions through a binary direction indicator.

\textbf{View Translation Definition.}  
Let $x_{{ref}}$, $x_{{tar}} \in \mathbb{R}^{3 \times H \times W}$ denote the reference and target view mammogram images, respectively.  
We introduce a binary direction indicator $d \in \{0, 1\}$ to control the synthesis direction:
\begin{equation}
    d=
  \begin{cases}
    0:  \text{CC} \rightarrow \text{MLO}, ~ x_{{ref}} = x_{{CC}}, ~ x_{{tar}} = x_{{MLO}}, \\
    1:  \text{MLO} \rightarrow \text{CC}, ~ x_{{ref}} = x_{{MLO}}, ~ x_{{tar}} = x_{{CC}},
  \end{cases}
  \label{eq:direction}
\end{equation}
where $x_{{CC}}$ and $x_{{MLO}}$ denote the CC and MLO view images, respectively.

To incorporate both the timestep $t$ and the translation direction $d$ into the diffusion process, we compute a learnable conditional embedding as the element-wise sum of the timestep embedding $t_{emb}$ and the direction embedding $d_{emb}$:
\begin{equation}
    {c}_{emb} = t_{emb} + d_{emb}.
\end{equation}
The conditional embedding ${c_{emb}}$ is then injected into each residual block of the UNet, allowing the model to modulate the denoising dynamics according to both temporal progression and the desired translation direction.

\textbf{Diffusion Forward Process.}
To facilitate efficient training and high-quality generation, we adopt a pretrained variational autoencoder (VAE)~\cite{vae} to encode input images into a latent representation \( z = \mathcal{E}(x) \in \mathbb{R}^{4 \times 32 \times 32} \), where \( \mathcal{E}(\cdot) \) denotes the VAE encoder.
Following the standard DDPM formulation, the forward diffusion process is simulated by progressively adding Gaussian noise to the target latent code. At each timestep $t\in [1,T]$, the noised target latent is computed as:
\begin{equation}
    z_{tar}^{t} = \sqrt{\bar{\alpha}_t} \mathcal{E}(x_{tar}) + \sqrt{1 - \bar{\alpha}_t} \epsilon ~~,\epsilon \sim \mathcal{N}(\textbf{0}, \textbf{I}), 
\end{equation}
where $\bar{\alpha}_t$ is the cumulative product of noise scheduling coefficients.

\textbf{Noise Prediction.}
The UNet-based denoising network $\epsilon_\theta$ is trained to predict the added noise $\epsilon$, conditioned on both the reference view $x_{ref}$ and the combined embedding $c_{emb}$. The objective function minimizes the mean squared error between the predicted and added noise: 
\begin{equation}
    \mathcal{L}=\mathbb{E}_{z_{\text {tar }}, \epsilon, t}\left\|\epsilon-\epsilon_{\theta}\left({z}_{tar}^{t}, c_{emb}, \mathcal{E}(x_{ref})\right)\right\|^{2}.
    \label{eq:oriloss}
\end{equation}

\subsection{Column-wise Correlation Decay Attention}
\begin{figure}
    \centering
    \includegraphics[width=\linewidth]{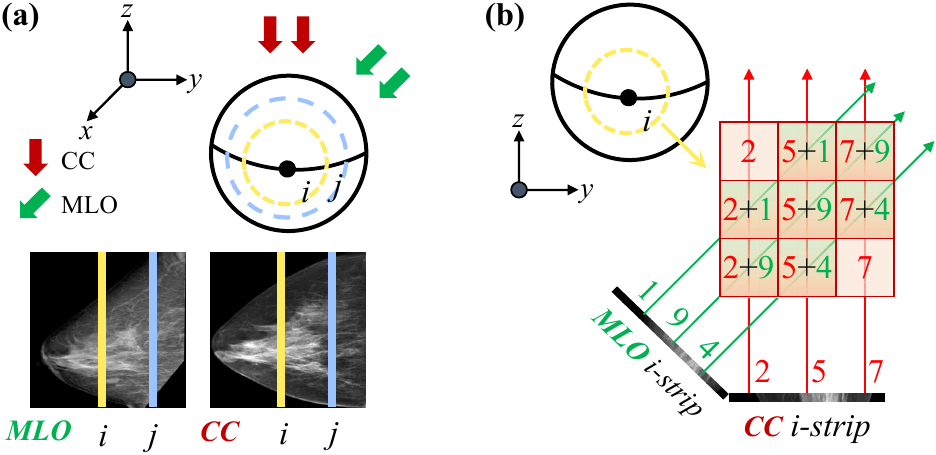}
    \caption{Schematic illustration of mammogram projection and back-projection.
    (a) The 3D breast volume is projected from two distinct viewpoints to produce the CC and MLO 2D mammogram images. The indices $i$ and $j$ indicate the positions of two 3D slices that appear as strip-like regions in the 2D projections.
    (b) The 3D slice is reconstructed by back-projecting the strip regions extracted from both views.}
    \label{fig:2}
\end{figure}

Unlike natural images, bilateral views in mammography capture different spatial projections of the same underlying breast volume. The anatomical structure of the breast can be approximately modeled as a rigid semi-spherical shape connected to the chest wall (Fig.~\ref{fig:2}a). Based on this assumption, 3D slices at equal distances from the chest wall are projected onto elongated regions in both the CC and MLO views. Although these projections are captured from different angles, they exhibit strong structural and semantic consistency.

This observation motivates a column-wise anatomical prior, wherein pixels aligned at the same horizontal position across the CC and MLO views are likely to encode similar tissue structures, while those farther apart exhibit reduced correlation. Accurately modeling and exploiting such localized correspondences is key to effective cross-view image translation.

\textbf{Cross-Attention in Latent Diffusion Models.}
In latent diffusion models, a common conditioning strategy involves reshaping a spatial feature map $f \in \mathbb{R}^{c \times h \times w}$ into a sequences of $N = h \times w$ tokens, each of dimension $c$, followed by the application of cross-attention~\cite{transformer} using a conditioning embedding. For example, in text-guided generation, the image tokens attend to text embeddings (\textit{e.g.}, from CLIP~\cite{CLIP}) via:
\begin{equation}
    \mathtt{Attn}(Q, K, V)=\operatorname{softmax}\left(\frac{Q K^{\top}}{\sqrt{d_{k}}}\right) V,
\end{equation}
where $Q$ is derived from the image features and $K$, $V$ are from the text condition. While effective for global control, this formulation does not explicitly account for spatial alignment, which is essential in medical imaging where anatomical coherence is critical.

\textbf{Column-Aware Cross-Attention (CACA) Mechanism.}
To better exploit anatomical consistency across views, we propose the CACA mechanism that leverages column-wise spatial alignment between the reference feature $f_{{ref}}$ and the target feature $f_{{tar}}$. As shown in Fig.~\ref{fig:1}b, we reshape both features into $N$ token sequences and compute:
\begin{equation}
Q = f_{{tar}} W_Q, \quad K = f_{{ref}} W_K, \quad V = f_{{ref}} W_V,
\end{equation}
where $W_Q$, $W_K$, and $W_V$ denote the linear projections for the queries, keys, and values, respectively.

We then inject a structured spatial prior into the attention weights using a Gaussian-decayed bias:
\begin{equation}
f_{CACA}=\mathtt{CACA}(Q, K, V) = \operatorname{softmax}\left(\frac{Q K^{\top}}{\sqrt{d_k}} + {col_{bias}} \right) V.
\end{equation}
The column-wise bias $col_{bias} \in \mathbb{R}^{N \times N}$ is defined as:
\begin{equation}
    col_{bias}^{(i, j)}=-\frac{\left(\Delta_{ {col }}^{(i, j)}\right)^{2}}{2 \sigma^{2}}, \quad \Delta_{ {col }}^{(i, j)}=\left|\operatorname{col}_{i}-\operatorname{col}_{j}\right|,
\end{equation}
where $\operatorname{col}_i$ and $\operatorname{col}_j$ denote the column features at spatial positions $i$ and $j$, respectively, and $\sigma$ controls the rate of decay.

This Gaussian-decayed prior softly enforces stronger interactions between spatially aligned regions across views while suppressing attention to mismatched areas. By encouraging the alignment of semantically consistent features, the CACA mechanism enhances anatomical fidelity in cross-view translation. It plays a crucial role in mitigating geometric mismatches and guiding our CA3D-Diff to generate more coherent and clinically realistic mammograms.

\subsection{Implicit 3D Structure Reconstruction}
To enhance anatomical consistency in cross-view mammogram translation, we propose an implicit 3D structure reconstruction module that derives a coarse volumetric representation from 2D latent features. This reconstructed volume acts as a global anatomical prior, guiding the diffusion model toward anatomically coherent and realistic synthesis.

\textbf{3D Geometry of Mammogram Projections.} 
Mammographic views can be approximately modeled as orthographic projections of a 3D volume onto 2D planes (Fig.~\ref{fig:2}a). Let $(x,y,z) \in \mathbb{R}^3$ denote a point in the 3D breast anatomy. We define a projection matrix $\mathbf{P}$ and a rotation matrix $\mathbf{R}$ as:
\begin{equation}
    \mathbf{P}=\left[\begin{array}{ccc}
1 & 0 & 0 \\
0 & 1 & 0 \\
0 & 0 & 0
\end{array}\right], \quad
\mathbf{R}=\left[\begin{array}{ccc}
1 & 0 & 0 \\
0 & \cos \theta & -\sin \theta \\
0 & \sin \theta & \cos \theta
\end{array}\right],
\end{equation}
where $\theta=45^{\circ}$ represents the rotation angle to MLO view.

The CC view can be interpreted as an orthographic projection onto the $z=0$ plane:
\begin{equation}
    \left(x, y, z\right)_{CC}=\mathbf{P}(x, y, z)^\text{T}=(x,y,0).
\end{equation}
The MLO view is acquired from an oblique plane approximately aligned with $z+y=0$, which corresponds to a rotation of $\theta=45^{\circ}$ around the $x$-axis:
\begin{equation}
    \left(x, y, z\right)_{MLO}=\mathbf{P}[\mathbf{R}(x, y, z)^\text{T}]=(x,(y-z)/\sqrt2,0).
\end{equation}
In both cases, intensity values along the depth dimension are aggregated to obtain a 2D image.

Leveraging these geometric priors, we back-project the 2D latent features into a coarse 3D feature volume, as shown in Fig.~\ref{fig:2}b. Each feature from the CC and MLO latents is mapped to its corresponding 3D location based on the projection model, and the resulting tensors are aggregated into a volumetric latent representation.

\textbf{3D Feature Refinement and Injection.} 
The initial 3D feature volume is refined using a lightweight 3D convolutional network to enhance spatial continuity and suppress projection-induced noise, resulting in a smoother representation $f_{3D}$. To incorporate this anatomical prior into the 2D diffusion process, we inject $f_{3D}$ into the UNet through a learned attention mechanism (Fig.~\ref{fig:1}c):
\begin{equation}
    f_{tar}'=f_{CACA}+\operatorname{ZeroConv}(\operatorname{Attention}(f_{CACA},f_{3D},f_{3D})),
\end{equation}
where $f_{tar}'$ serves as the input to the next CACA module. The attention module aligns the current feature map with the 3D structural context, while the ZeroConv layer (a $1\times1$ convolution with zero-initialized weights) enables the network to gradually incorporate 3D information without disrupting early training dynamics.

\subsection{Training and Inference Strategy}
During training, both the reference and target view images are encoded into latent representations $z_{ref}$ and $z_{tar}$ via VAE encoder. Gaussian noise is then added through the forward diffusion process. The resulting noisy latents are back-projected to construct the 3D feature volume $f_{3D}$, which guides the denoising UNet to recover $z_{tar}^t$. 

For inference, the target latent $z_{tar}^{T}$ is initialized from $\mathcal{N}(\textbf{0}, \textbf{I})$. The reference view is encoded via VAE encoder and noised to a synchronized timestep $t$, enabling the construction of the corresponding 3D feature volume $f_{3D}$. The volume then guides the iterative denoising process from timestep $T$ to 0. Finally, the denoised latent is decoded by VAE decoder to generate the synthesized target view image $\hat{x}_{tar}$.

\section{Experiments}
\subsection{Materials and Details}
\paragraph{Dataset}
All experiments are conducted on the publicly available VinDr-Mammo dataset~\cite{dataset}. We extract paired craniocaudal (CC) and mediolateral oblique (MLO) views of the same breast, yielding 6,700 pairs for training, 748 pairs for validation, and 1,860 pairs for testing.
To preprocess the data, we employ the OpenBreast library~\cite{openbreast} to generate breast segmentation masks, which are subsequently used to remove non-breast background regions and pectoralis muscles. To ensure consistent anatomical alignment, all images are shifted such that the chest wall aligns with the right edge of the image. Each image is resized to $256\times 256$, followed by truncation normalization to enhance contrast and facilitate model training.


\paragraph{Evaluation Metrics}
To evaluate the quality of synthesized mammogram views, we adopt both pixel-level and perceptual-level metrics. Specifically, we use Structural Similarity Index Measure (SSIM) and Peak Signal-to-Noise Ratio (PSNR)~\cite{psnrvsssim} to quantify low-level similarity between the generated images and ground truth, reflecting structural and photometric fidelity.
In addition, we evaluate perceptual realism using Fréchet Inception Distance (FID)~\cite{Seitzer2020FID} and Learned Perceptual Image Patch Similarity (LPIPS)~\cite{lpips}, which measure high-level feature similarity. FID is computed based on deep features extracted by a pre-trained Inception-V3 network~\cite{inception}, while LPIPS is computed using the VGG16 model~\cite{vgg} as the backbone. These two metrics assess the closeness between the distributions of generated and real images from a perceptual perspective, and are widely used for evaluating image synthesis quality.

\paragraph{Implementation Details}
Our CA3D-Diff framework is implemented in PyTorch and trained on a single NVIDIA A100 80GB GPU. The diffusion process is configured with $T=1000$ steps, and the noise schedule $\{\beta_i\}^T_{i=1}$ is linearly interpolated from $8.5e-4$ to $0.012$. We adopt the AdamW optimizer with an initial learning rate of $1e-4$, combined with a Lambda Linear learning rate scheduler. Training is conducted for 40,000 steps (approximately 23 hours) using a batch size of 32, with a GPU memory footprint of around 33GB.
During inference, we use 50 denoising steps, which takes approximately 7.5 seconds and 4GB of memory per sample. The CACA module applies a Gaussian decay prior with $\sigma=5$ to modulate cross-view feature similarity based on spatial distance. To improve generation fidelity, we employ classifier-free guidance~\cite{CFG}, where conditional inputs are randomly masked with a probability of 0.1 during training, and a guidance scale of 3.0 is applied at inference.

\begin{figure*}
    \centering
    \includegraphics[width=0.9\linewidth]{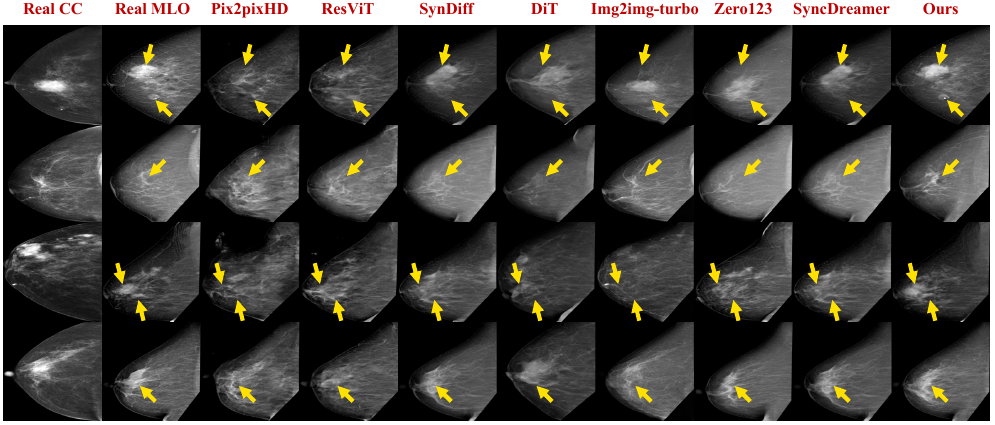}
    \caption{Qualitative comparison on the VinDr-Mammo dataset for $CC \rightarrow MLO$. The first and second columns show the input CC and ground-truth MLO images, respectively. Yellow arrows mark anatomical differences, with our CA3D-Diff producing more accurate results compared to other SOTA methods.}
    \label{fig:3}
\end{figure*}

\begin{figure*}
    \centering
    \includegraphics[width=0.9\linewidth]{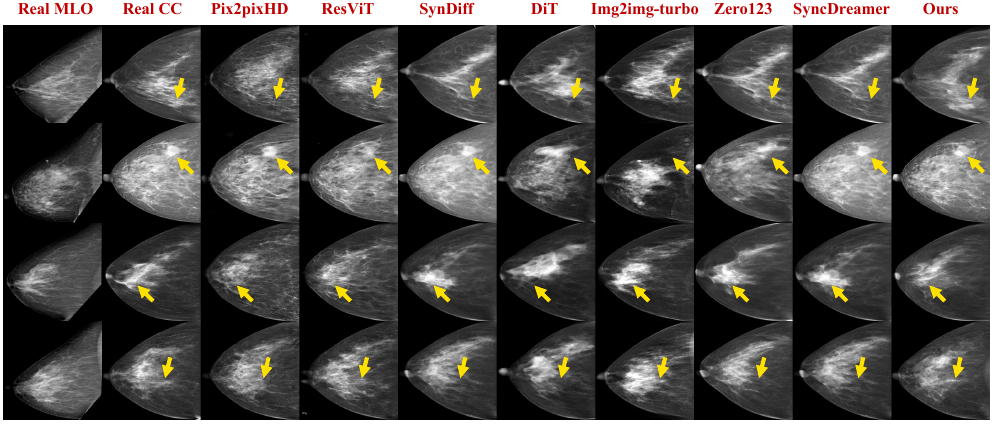}
    \caption{Qualitative results on the VinDr-Mammo dataset for $MLO \rightarrow CC$ translation. The first column shows input MLO images, and the second column presents the ground-truth CC images. Yellow arrows mark anatomical differences, where our method demonstrates improved structural fidelity.}
    \label{fig:4}
\end{figure*}

\subsection{Comparison with SOTA Methods}\label{AA}

\paragraph{Quantitative Comparison}

\begin{table}
\centering
\caption{Quantitative comparison with existing methods on VinDr-Mammo dataset. \textbf{Bold}: Best result. \underline{Underline}: Second-best result.}
\label{tab:1}
\resizebox{\columnwidth}{!}{
\begin{tblr}{
  cells = {c},
  cell{2}{1} = {r=9}{},
  cell{11}{1} = {r=9}{},
  vline{2-3} = {-}{},
  hline{2,11} = {-}{},
  hline{1,20} = {-}{1.5pt},
}
\textbf{{Direction}} & \textbf{{Methods}} & \textbf{\textit{PSNR$\uparrow$}} & \textbf{\textit{SSIM$\uparrow$}} & \textbf{\textit{FID$\downarrow$}} & \textbf{\textit{LPIPS$\downarrow$}} \\
CC$\rightarrow$MLO           & CycleGAN                  & 17.013                  & 0.464                   & 35.803                 & 0.307                    \\
                            & Pix2pixHD                 & 17.830                  & 0.476                   & 39.199                 & 0.301                    \\
                            & ResViT                    & \underline{18.408}          & 0.487                   & 36.097                 & 0.288                   \\
                            & SynDiff                   & 16.841                  & 0.463                   & 41.267                 & 0.318 \\
                            & DiT                       & 17.378                  & 0.493                   & 27.670                 & 0.302
                            \\
                            & Img2img-turbo             & 16.804                  & 0.452                   & 34.237                 & 0.312                    \\
                            & Zero123                   & 17.747                  & 0.493                   & 19.193                 & 0.291                   \\
                            & SyncDreamer               & 18.219                  & ~\underline{0.506}          & \underline{17.657}         & \underline{0.284}                    \\
                            & CA3D-Diff (Ours)                      & \textbf{20.537}         & \textbf{0.590}          & \textbf{15.840}        & \textbf{0.214}                          \\
MLO$\rightarrow$CC                     & CycleGAN                  & 15.342                  & 0.392                   & 42.501                 & 0.316                    \\
                            & Pix2pixHD                 & 16.724                  & 0.404                   & 44.526                 & 0.305                    \\
                            & ResViT                    & \underline{16.821}          & 0.408                   & 31.678                 & 0.302                    \\
                            & SynDiff                   & 15.635                  & 0.368                   & 41.842                 & 0.328                \\
                            & DiT                       & 15.699                  & 0.416                   & 21.900                 & 0.314                \\
                            & Img2img-turbo             & 15.010                  & 0.386                   & 26.815                 & 0.328                    \\
                            & Zero123                   & 15.974                  & 0.417                   & 18.564                 & 0.303                    \\
                            & SyncDreamer               & 16.291                  & \underline{0.427}           & \textbf{16.434}        & \underline{0.299}            \\
                            & CA3D-Diff (Ours)                      & \textbf{17.708}         & \textbf{0.464}          & \underline{17.461}         & \textbf{0.268}                
\end{tblr}
}
\end{table}

We conduct quantitative comparisons against eight representative image translation and synthesis baselines, including GAN-based models (CycleGAN~\cite{cyclegan}, Pix2pixHD~\cite{pix2pixhd}, ResViT~\cite{resvit}) and diffusion-based models (DiT~\cite{dit}, SynDiff~\cite{syndiff}, Zero123~\cite{zero123}, Img2img-Turbo~\cite{img2img-turbo}, SyncDreamer~\cite{syncdreamer}). All methods are implemented using their official open-source code and trained on the same train-test split paired mammogram dataset. For single-direction models, we train two separate networks to support bidirectional view translation.
To tailor the DiT for our task, we discard its original class label conditioning and replace it with conditioning on the reference image, implemented via feature concatenation.
Furthermore, we remove camera parameter inputs in Zero123 and SyncDreamer and provide 3D features reconstructed based on mammogram projection geometry. 

As shown in Table~\ref{tab:1}, CA3D-Diff consistently outperforms all state-of-the-art methods across both $CC\rightarrow MLO$ and $MLO\rightarrow CC$ tasks. In the $CC\rightarrow MLO$ direction, it achieves over 11\% improvement in PSNR, nearly 17\% in SSIM, and reduces FID and LPIPS by more than 10\% and 24\%, respectively. In the reverse direction, it still outperforms the second-best method with relative gains exceeding 5\% in PSNR, 8\% in SSIM, and a 10\% reduction in LPIPS, while maintaining competitive FID. These results confirm the effectiveness of CA3D-Diff in generating perceptually and structurally faithful cross-view translations.

\paragraph{Qualitative Comparison}

We conduct a visual comparison of our CA3D-Diff with several SOTA methods on the VinDr-Mammo dataset.
Fig.~\ref{fig:3} and Fig.~\ref{fig:4} present representative examples of $CC\rightarrow MLO$ and $MLO\rightarrow CC$ translation, respectively. The first column shows the input reference images, while the second column displays the corresponding ground-truth target views. The remaining columns depict the results generated by different methods.
CA3D-Diff more effectively preserves anatomical structures and fine-grained textures, particularly in complex tissue regions, compared to existing baselines, as highlighted by the yellow arrows. These results highlight the strength of our method in producing anatomically consistent and realistic cross-view mammograms.

\subsection{Ablation Study}
\begin{table}
\centering
\caption{Ablation study of the proposed components in CA3D-Diff on VinDr-Mammo Dataset. CACA and IM3D represent column-aware cross-attention mechanism and implicit 3D structure reconstruction module, respectively.}
\label{tab:2}
\resizebox{\columnwidth}{!}{
\begin{tblr}{
  cells = {c},
  cell{2}{1} = {r=4}{},
  cell{6}{1} = {r=4}{},
  vline{2,4} = {-}{},
  hline{2,6} = {-}{},
  hline{1,10} = {-}{1.5pt},
}
\textbf{{Direction}} & \textbf{{CACA}} & \textbf{{IM3D}} & \textbf{\textit{PSNR$\uparrow$}} & \textbf{\textit{SSIM$\uparrow$}} & \textbf{\textit{FID$\downarrow$}} & \textbf{\textit{LPIPS$\downarrow$}} \\
CC$\rightarrow$MLO          & \xmark & \xmark & 17.107 & 0.477 & 28.810 & 0.268 \\
                            & \cmark & \xmark & 18.326 & 0.524 & 20.932 & 0.232 \\
                            & \xmark & \cmark & 18.419 & 0.520 & 21.164 & 0.234 \\
                            & \cmark & \cmark & \textbf{20.537} & \textbf{0.590} & \textbf{15.840} & \textbf{0.214} \\
MLO$\rightarrow$CC          & \xmark & \xmark & 15.322 & 0.409 & 24.273 & 0.320 \\
                            & \cmark & \xmark & 16.321 & 0.444 & 19.541 & 0.288 \\
                            & \xmark & \cmark & 16.354 & 0.440 & 19.823 & 0.291 \\
                            & \cmark & \cmark & \textbf{17.708} & \textbf{0.464} & \textbf{17.461} & \textbf{0.268} \\
\end{tblr}
}
\end{table}

To further investigate the contribution of each module in CA3D-Diff, we conduct an ablation study by incrementally adding the column-aware cross-attention (CACA) and implicit 3D structure reconstruction (IM3D) modules. The results are summarized in Table~\ref{tab:2}, covering both translation directions.

Starting from a baseline model, we observe notable performance gains when incorporating CACA, which enhances spatial feature alignment across views. Similarly, adding IM3D alone also brings comparable improvements by introducing 3D-aware feature modeling, helping the network reconstruct anatomically plausible targets even under complex deformations.
When both CACA and IM3D are integrated, the model achieves the best performance across all metrics. For example, in the $CC\rightarrow MLO$ direction, PSNR increases from $17.107$ to $20.537$, and FID drops from $28.810$ to $15.840$. Similar trends are observed in the $MLO\rightarrow CC$ direction.

These results demonstrate the complementary roles of the two modules: CACA improves cross-view attention and structural alignment, while IM3D contributes to global anatomical consistency and visual realism. Their combination leads to substantial performance gains, validating the effectiveness of the full CA3D-Diff architecture.

\subsection{Clinical Application in Screening}
To assess the clinical utility of our proposed CA3D-Diff framework, we conduct a downstream benign vs. malignant classification task on the mass subset of the VinDr-Mammo dataset, focusing on classification with view augmentation. 

Following standard clinical practice, we group BI-RADS categories 1–3 as benign and categories 4–5 as malignant. The training set comprises 191 benign and 227 malignant cases, while the test set includes 57 benign and 45 malignant samples. We use a ResNet-50~\cite{resnet} classifier and report three commonly adopted clinical metrics: accuracy, AUC, and F1-score.

To examine the benefit of synthesized cross-view images, we compare five input settings: using a single real CC or MLO view, combining one real view with its corresponding synthesized counterpart, and using both real CC and MLO views. For dual-view inputs, images are concatenated along the channel dimension. During both training and inference, synthesized views are consistently used in the augmented settings. The real CC + real MLO setup serves as the upper performance bound.

\begin{table}
\centering
\caption{Benign vs. malignant classification results on the mass subset of VinDr-Mammo using ResNet-50. ``Synth MLO'' and ``Synth CC'' denote cross-view images generated by CA3D-Diff from real CC and MLO inputs, respectively.}
\label{tab:3}
\resizebox{\columnwidth}{!}{
\begin{tblr}{
  cells = {c},
  row{3,4} = {bg=blue!8},
  row{5,6} = {bg=green!8},
  vline{2} = {-}{},
  hline{1,7} = {-}{1.5pt},
  hline{2} = {-}{},
  hline{3} = {-}{dashed},
}
\textbf{Data Setting}      & \textbf{\textit{Accuracy}}~(\%)$~\uparrow$ & \textbf{\textit{AUC}}~(\%)$~\uparrow$ & \textbf{\textit{F1-score}}~(\%)$~\uparrow$ \\
Real CC + Real MLO      & 74.5 & 73.2 & 68.3      \\
Real CC            & 67.3  & 66.8 & 63.0   \\
Real CC + Synth MLO     & $72.5_{\color{red}+5.2}$ & $71.9_{\color{red}+5.1}$ & $68.2_{\color{red}+5.2}$   \\
Real MLO           & 68.6  & 67.7 & 62.8   \\
Synth CC + Real MLO     & $73.5_{\color{red}+4.9}$  & $72.6_{\color{red}+4.9}$ & $68.2_{\color{red}+5.4}$  
\end{tblr}
}
\end{table}

As shown in Table~\ref{tab:3}, incorporating synthesized cross-view images consistently improves classification performance over single-view baselines across all metrics. The MLO view consistently yields slightly better classification performance compared to the CC view, which aligns with clinical practice where MLO is often prioritized for preliminary diagnosis due to its higher diagnostic information density. Notably, using one real and one synthesized view achieves performance close to the dual-real-view setup, suggesting that our generated views retain essential diagnostic information.
These results demonstrate the practical value of CA3D-Diff in screening scenarios with limited view availability, underscoring its potential as an effective data augmentation strategy in clinical workflows.

\section{Conclusion}

We present CA3D-Diff, a novel conditional diffusion framework that tackles the challenging task of bidirectional mammogram view translation between cranio-caudal (CC) and mediolateral oblique (MLO) views. By incorporating a column-aware cross-attention mechanism that exploits the anatomical alignment along vertical columns, and an implicit 3D structure reconstruction module that approximates the underlying breast structure through projection geometry, our method effectively models the complex, non-rigid cross-view mapping without requiring pixel-wise correspondence or explicit 3D supervision. Extensive experiments on the VinDr-Mammo dataset demonstrate that CA3D-Diff consistently outperforms existing baselines in both visual fidelity and structural consistency. 
Furthermore, the synthesized views provide practical benefits for downstream screening tasks, underscoring the clinical relevance of our approach. 
This work offers new insights into anatomically guided mammography synthesis, with potential applications in missing view recovery, data augmentation, and cross-view representation learning for computer-aided diagnosis systems.

\bibliographystyle{IEEEtran}
\balance
\bibliography{Ref.bib}

\end{document}